# Spatiality–Frequency Domain Video Forgery Detection System Based on ResNet-LSTM-CBAM and DCT Hybrid Network

**Zihao Liao[1], Sheng Hong[2,3],* and Yu Chen [1]**

[1] School of Information Engineering, Nanchang University, Nanchang 330031, China; zhliao@email.ncu.edu.cn (Z.L.); ychen@email.ncu.edu.cn (Y.C.)
[2] Nanchang University, Nanchang 330031, China
[3] School of Cyber Science and Technology, Beihang University, Beijing 100191, China
* Correspondence: hongsheng@ncu.edu.cn

**Abstract**

As information technology advances, digital content has become widely adopted across diverse fields such as news broadcasting, entertainment, commerce, and forensic investigation. However, the availability of sophisticated multimedia editing tools has significantly increased the risk of video and image forgery, raising serious concerns about content authenticity at both societal and individual levels. To address the growing need for robust and accurate detection methods, this study proposes a novel video forgery detection model that integrates both spatial and frequency-domain features. The model is built on a ResNet-LSTM framework enhanced by a Convolutional Block Attention Module (CBAM) for spatial feature extraction, and further incorporates Discrete Cosine Transform (DCT) to capture frequency domain information. Comprehensive experiments were conducted on several mainstream benchmark datasets, encompassing a wide range of forgery scenarios. The results demonstrate that the proposed model achieves superior performance in distinguishing between authentic and manipulated videos. Additional ablation and comparative studies confirm the contribution of each component in the architecture, offering deeper insight into the model's capacity. Overall, the findings support the proposed approach as a promising solution for enhancing the reliability of video authenticity analysis under complex conditions.

## 1. Start

### *1.1. Introduction*

The rapid advancement of the Internet has triggered a series of data security crises, posing significant threats to information safety across various sectors. As a core component of modern infrastructure, national information services demand robust data security mechanisms to ensure stable and trustworthy operations. In response to these challenges, governments around the world have issued laws and regulations aimed at guiding the development of information security frameworks and establishing explicit requirements for data protection.

For instance, in the domain of traffic data security, Hong et al. [1] proposed resilience-based strategies for mitigating cyberattacks on critical infrastructure, emphasizing fault propagation modeling and trend forecasting to support adaptive recovery. Meanwhile, with the proliferation of multimedia technologies and online platforms, digital images and videos have become pervasive in daily life, serving as major carriers of information. However, powerful editing tools—particularly those driven by artificial intelligence—have made it increasingly easy to manipulate digital content, leading to the widespread emergence of deepfakes and other forms of tampered media.

Such manipulated content, when maliciously disseminated, may pose serious risks. It may mislead the public with false information, distort societal values, erode legal and political trust, and even incite conflicts or widespread panic. In high-risk domains such as judicial procedures, journalism, political communication, and financial transactions, forged videos can severely damage the authenticity and integrity of evidence or information. Therefore, the development of an automated and reliable video forgery detection system has become a technological imperative and a societal necessity. We need such a system to identify forged videos, thereby locating manipulated areas, analyzing types of forgery, and ultimately verifying the credibility of content in an environment increasingly influenced by artificial intelligence.

Existing video forgery detection methods, although increasingly sophisticated, still face notable challenges. Many deep learning-based models are confined to a single feature domain—such as spatial, temporal, or frequency—and therefore struggle to generalize across diverse types of forgery. Moreover, most existing approaches are tailored to specific attack scenarios (e.g., face swapping or frame deletion), making them less adaptable to emerging forgery paradigms, particularly those generated by AIGC (AI-generated content). In addition, the limited availability of annotated training data and the inherent complexity of multimodal video sequences often lead to overfitting and poor generalization performance.

To address these challenges, this study makes the following contributions:

First, we propose a hybrid deep learning model that jointly leverages spatial and frequency domain features for improved detection accuracy. Our architecture is based on the ResNet-LSTM-Attention framework, further enhanced by a Convolutional Block Attention Module (CBAM) to capture both spatial and channel-wise dependencies. In addition, we incorporate the Discrete Cosine Transform (DCT) to extract the frequency domain representations, enabling the model to perform complementary domain analysis for more robust forgery detection.

Second, our method is capable of detecting various types of video forgery—including frame deletion, insertion, replacement, and manipulations generated by AI-generated content (AIGC).

Third, although the CNN–LSTM–Attention architecture [2–6] has become a mainstream paradigm in video forgery detection due to its capacity to model spatial dependencies, it still faces several challenges. Specifically, the architeture tends to overfit on complex multimodal sequences, particularly under conditions of limited and imbalanced training data. To overcome these limitations, we propose a hybrid optimization strategy that builds upon the CNN–LSTM–Attention framework while incorporating structural and training-level adjustments designed to enhance generalization, increase robustness against data imbalance, and mitigate overfitting when dealing with diverse and complex forgery patterns.

### *1.2. Literature Review*

Prior to the emergence of deep learning, a variety of traditional methods were developed to tackle video forgery detection. These approaches typically relied on handcrafted features and statistical analysis, and can be broadly grouped into four main categories. Pixel-level correlation approaches, such as those proposed by Sitara et al. [7] and Liu et al. [8], detect anomalies based on differences in pixel values and inter-frame variations including brightness and contrast. Statistical feature-based methods were also prevalent; for example, Saddique et al. [9] employed an edge image-based likelihood ratio approach, while Chen et al. [10] introduced a variable width boundary region algorithm to assess frame completeness. Other researchers focused on noise pattern-based techniques: Kobayashi et al. [11] investigated the spatial distribution of sensor noise, and Goodwin and Chetty [12] proposed a sub-block level observer that combines noise and quantization residue analysis. Lastly, content-based detection methods interpret semantic and motion information. For instance, D'Amiano et al. [13] applied block-matching algorithms to localize tampered areas, while Aloraini et al. [14] attempted to reverse estimate deleted object trajectories. Although these methods offered interpretable and sometimes effective solutions, they often suffered from poor generalizability, sensitivity to parameter settings, and limited robustness when faced with complex, real-world tampering scenarios.

With the rise of deep learning, researchers began to explore models that automatically extract and learn discriminative features from data. Spatial–temporal detection models have become widely adopted due to their ability to simultaneously capture spatial information and motion continuity. For instance, the CST-FDNet model [15] introduced a cross-layer attention mechanism that improves the localization accuracy of tampered regions. Meanwhile, some approaches have been inspired by biological signals. Fei Siyu [16] incorporated pupil features and multimodal fusion strategies to detect forged face videos, and Liang Jiahao [17] leveraged physiological signals such as remote photoplethysmography (rPPG) to enhance detection accuracy and robustness. Lu Wei's group at Sun Yat-Sen University proposed the CFPRF framework [18], which is designed for detecting minor tampering in video sequences. Their model integrates a difference-aware feature-learning module with a boundary-enhancing module, using contrastive representation learning to magnify subtle discrepancies between authentic and manipulated frames. These methods mark a clear transition from conventional feature engineering to deep representation learning, significantly improving forgery detection performance.

To address increasingly complex manipulation types, especially those generated by AI-generated content (AIGC), several recent works have begun to incorporate multimodal information and large-scale pre-trained models. The Sora Detector system [19] exemplifies this trend by combining keyframe extraction with large-scale visual language models and knowledge graphs for enhanced generality. Other studies, such as those by DuB3D [20] and DeCoF [21], focus on temporal artifact detection and optical flow analysis, improving temporal consistency modeling between frames. Additionally, detection frameworks leveraging both local and global motion features have emerged [22], using channel attention to improve discrimination. More recently, large language models have been introduced into the video forensics domain; MM-Det [23] and Statedful-CCSH [24] represent early efforts to integrate semantic-level reasoning with temporal forgery detection, offering a new direction for addressing AIGC-related threats.

Despite these advances, the current research addresses several critical limitations. Most existing approaches rely heavily on spatial, temporal, or frequency domain features in isolation, which may not be sufficiently robust against sophisticated forgery techniques that exploit multiple modalities simultaneously. Furthermore, many models are designed for specific forgery types, such as face swapping or frame insertion, and therefore lack generalizability across unknown or novel manipulation forms. The limited availability of balanced, annotated datasets and the complex nature of multimodal video data also contribute to issues of overfitting and performance degradtion in real-world scenarios. These challenges motivate the development of hybrid models that integrate diverse feature domains and incorporate architectural designs capable of generalizing to novel forgery patterns—especially those stemming from rapidly evolving AIGC media.

## 2. Material and Methods

The datasets in this work are obtained from the Video Forgery Detection Database VFDD2.1 [25,26], which is published by South China University of Technology and the DVF dataset [23]. The VFDD2.1 database has 2189 videos in total, composed of 1496 normal videos, 496 inter-frame forged videos, and 197 intra-frame forged videos. On the other hand, the DVF dataset comprises many AI-generated fake videos that are made based on existing AIGC (Artificial Intelligence-Generated Content) tools using different sources of real videos. In this work, most of the dataset is built using videos from the DVF dataset and the rest used VFDD2.1. It consists of 1968 forged and 2329 real videos totaling 4297 video clips. Among these, 80 percent are taken for training and the other 20 percent are taken for validation.

More specifically, the main part of this dataset is derived from the DVF dataset, but some data from the VFDD2.1 dataset was selected during the construction process. Generally speaking, this is done to incorporate some traditional forgery methods into the model for training. As a result, it will not only have a certain ability to recognize AIGC forgeries, but also a certain ability to recognize traditional forgeries. In the VFDD2.1 dataset, three datasets of traditional tampering methods were selected for overall construction, and a portion was reserved for verification. As for the issue of class balance, in the VFDD2.1 dataset, there are multiple scenarios, so one real and two false ones were chosen to construct it. This is because in the DVF dataset, the real dataset comes from social software, while the false dataset comes from AI-generated content. Overall, the real dataset is more abundant, while the false dataset is less so. Therefore, in order to achieve a balance in dataset construction, some false datasets were introduced in the construction of the dataset. This is the selection of VFDD2.1. Finally, the ratio of true values to false values in the dataset did not reach a 1:1 ratio, so this situation was also taken into account when constructing the loss function. Overall, the Focal Loss function was used to modulate the categories of the original samples. The specific content of Focal Loss is described in Section 2.1.

The preprocessing pipeline consists of two main modules: label generation and video frame extraction. In the label generation module, the dataset directory is recursively scanned to automatically assign a label to each video clip. Videos are categorized as either authentic or forged. This process constructs binary classification labels by reading the video files in the dataset directory.

For video frame extraction, since the source videos have inconsistent input formats and different input sizes, a fixed-length sequence is extracted to select frames from each original video. Given that all the videos are relatively short, we set the input sequence length to 30 frames to meet the requirement of the deep learning model. Each video is decoded frame by frame, and all the frames are resized to 224 × 224 pixels to facilitate efficient feature extraction and improve training speed. If a video contains fewer than 30 frames, the last frame is duplicated to pad the sequence. If a video contains more than 30 frames, keyframes are uniformly sampled. The generated frame sequence is then reshaped into a (30, 3, 224, 224) tensor, where 30 is the number of frames, 3 is the number of RGB channels, and 224 × 224 is the spatial resolution.

To further enhance the model, various data augmentations are applied to the input data, including random horizontal flipping, random rotation, color jittering, random cropping, and random affine transformations. These augmentations transform the data in multiple ways, increasing the diversity of the training data distribution and improving the model's robustness to noise and perturbations.

In order to achieve faster and more stable training of the deep learning model, data normalization must be applied. Common normalization techniques include Min–Max normalization and Z-score normalization. Based on the characteristics of the spatial and frequency domain information, it is essential to determine an appropriate normalization stage. Spatial domain data are normalized during the frame extraction stage, while for frequency domain features extracted via the Discrete Cosine Transform (DCT), normalization is performed after the corresponding DCT coefficients are computed. In this study, we adopt Z-score normalization, which is defined as follows:

$$x_{\text{norm}} = \frac{x - \mu}{\sigma + \epsilon} \tag{1}$$

where $x$ is the original data value, $\mu$ is the mean of the dataset, $\sigma$ is the standard deviation, and $\varepsilon = 10^{-6}$ is a small constant added to avoid division by zero.

### 2.1. Overall Framework of the Proposed Algorithm

In this study, the proposed model employs a convolutional architecture based on ResNet-50, integrated with the Convolutional Block Attention Module (CBAM), to extract deep features from video frame sequences. The convolutional layers capture intra-frame spatial features and local texture information. Specifically, CBAM consists of two consecutive sub-modules: channel attention and spatial attention. The channel attention module infers a one-dimensional channel attention map (according to Formula (7)) by applying global average pooling and Max pooling to the input feature map, followed by a shared multi-layer perceptron (MLP). Next, the spatial attention module focusing on "where" needs to be emphasized or suppressed. For the specific calculation process, refer to Formula (9). Then, the extracted features are subsequently fed into an LSTM module, which models the temporal dependencies within the video sequence and captures the frame-to-frame dynamics as well as long-range contextual relationships. To enhance the discriminative power of feature extraction, an attention module is introduced into the model to dynamically learn the weight of each feature, assigning higher weights to features that contribute more to tampering trace detection. Although the model performs well in feature extraction, it can still be affected by complex backgrounds and local deformations.

To further improve model optimization and detection accuracy, the proposed design incorporates a DCT-based feature extraction module. Using Formula (14), Discrete Cosine Transform (DCT) features effectively represent the frequency domain characteristics of video frames. When these frequency domain features are combined with the spatial features extracted by ResNet-50 integrated with CBAM, the model can more effectively capture subtle patterns and fine-grained differences within video content.

The loss function used in the proposed model is Focal Loss. Focal Loss adaptively adjusts sample weights to alleviate class imbalance in the training set and enhance the model's ability to detect minority classes. Focal Loss is an enhancement of the cross-entropy loss function, incorporating a class-balancing parameter $\alpha = 0.5$ and a modulation factor $g = 0.2$. The overall loss of the model is computed as the mean Focal Loss across all samples. In the Adam optimizer, weight decay is applied through an L2 regularization term, which serves as an implicit loss to constrain the model's complexity. The details of Focal Loss are presented as follows:

Firstly, the formula for cross-entropy loss is as follows:

$$CE(p, y) = -\sum_{i=1}^{C} y_i \log(p_i) \tag{2}$$

$C$ represents the number of classes (for the video forgery detection task $C = 2$, indicating binary classification), $y_i$ is the true label, and $p_i$ is the predicted probability for class $i$.

For a single sample, the cross-entropy loss simplifies to

$$CE\,(p, y) = -\log(p_y) \tag{3}$$

$p_y$ represents the predicted probability of the true class $y$.

Eventually, we can obtain the expression of focal loss as follows:

$$FL\,(p_t) = -\alpha(1 - p_t)^{g} \log(p_t) \tag{4}$$

$p_t = p_y$ (the predicted probability for the true class), i.e., $p_t = p$ if $y_i = 1$, otherwise $p_t = 1 - p$. The class weight $\alpha = 0.5$ is used to balance the positive and negative samples, and the modulation factor $g = 0.2$ increases the focus on difficult samples.

The overall loss of the model is computed as the average Focal Loss across all samples, where N denotes the total number of samples. The specific formula is given as follows:

$$L = \frac{1}{N}\sum_{i=1}^{N} FL(p_t^{(i)}) \tag{5}$$

Furthermore, to accelerate convergence and prevent overfitting, the training process incorporates an adaptive learning rate strategy based on validation performance, along with a learning rate warm-up scheme. The final output of our forgery detection system is a binary decision indicating whether a video is forged or authentic.

The overall architecture of the proposed model is illustrated in Figure 1:

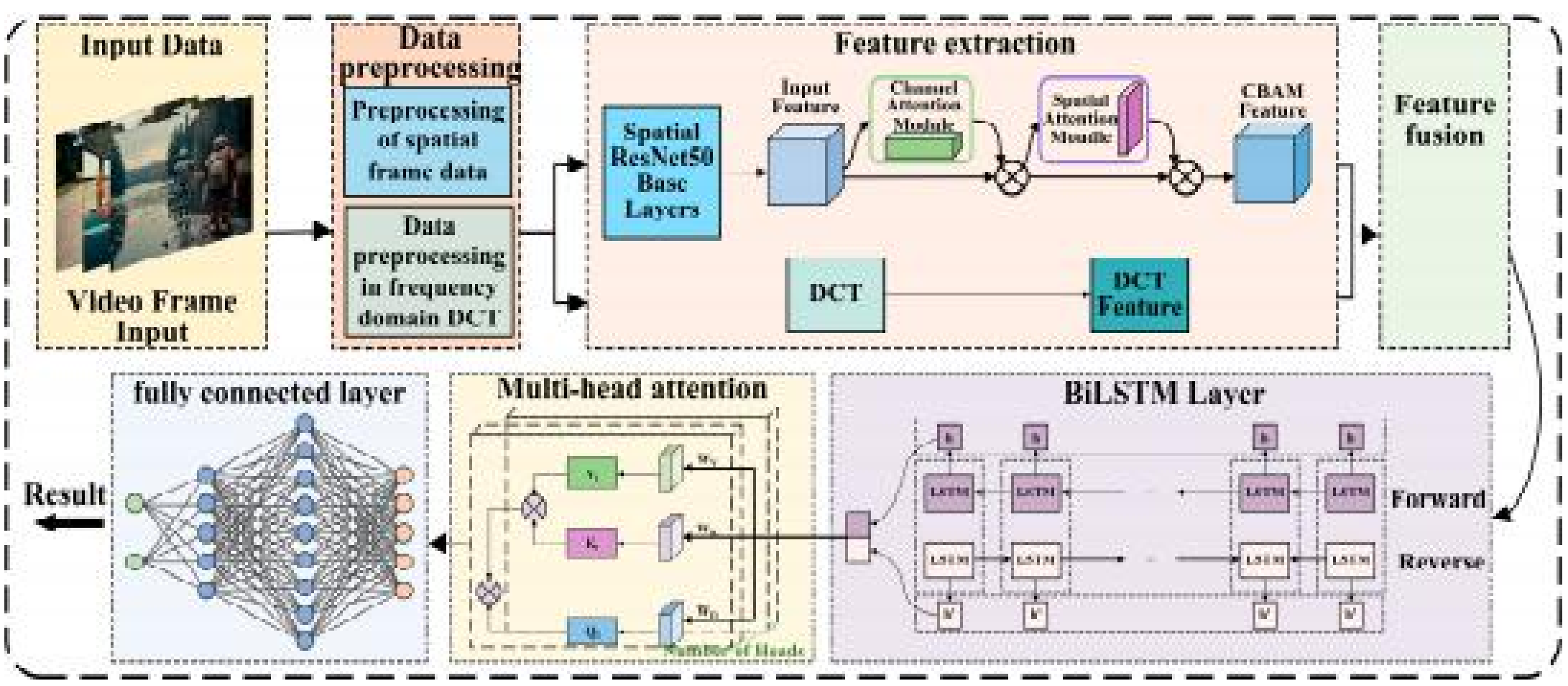

**Figure 1.** The overall structure diagram of our model.

*2.2. Integration of CBAM*

The Convolutional Block Attention Module (CBAM) [27] is a lightweight attention mechanism that can be seamlessly integrated into convolutional neural networks to enhance their performance. By leveraging attention mechanisms, the model is able to more effectively capture salient features [28]. CBAM applies attention along both the channel and spatial dimensions, explicitly modeling the importance of different channels and spatial locations within the feature map. This enables the network to focus more precisely on informative regions and enhances its representational capacity.

The core concept of CBAM is to process the input feature map through two sequential stages. First, channel attention is applied to emphasize what features are important, followed by spatial attention, which focuses on where the crucial information is located.

This dual-attention mechanism enables CBAM to extract key information from both the channel and spatial dimensions of the feature representation. The architecture of the CBAM module, including its dual-attention mechanism, is illustrated in Figure 2:

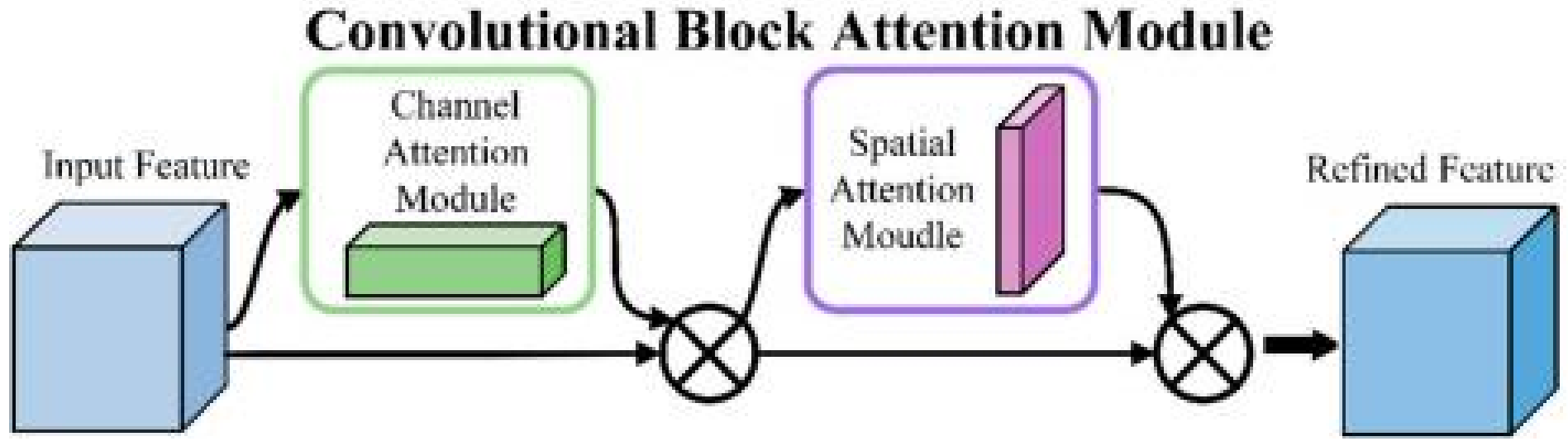

**Figure 2.** Schematic diagram of the dual attention mechanism module of CBAM.

CBAM is composed of two primary components: the Channel Attention Module and the Spatial Attention Module. The following sections provide a detailed explanation of each.

As illustrated in Figure 3, the Channel Attention Module is designed to emphasize the most informative feature channels while suppressing those that are less relevant. Its mechanism of operation can be summarized as follows:

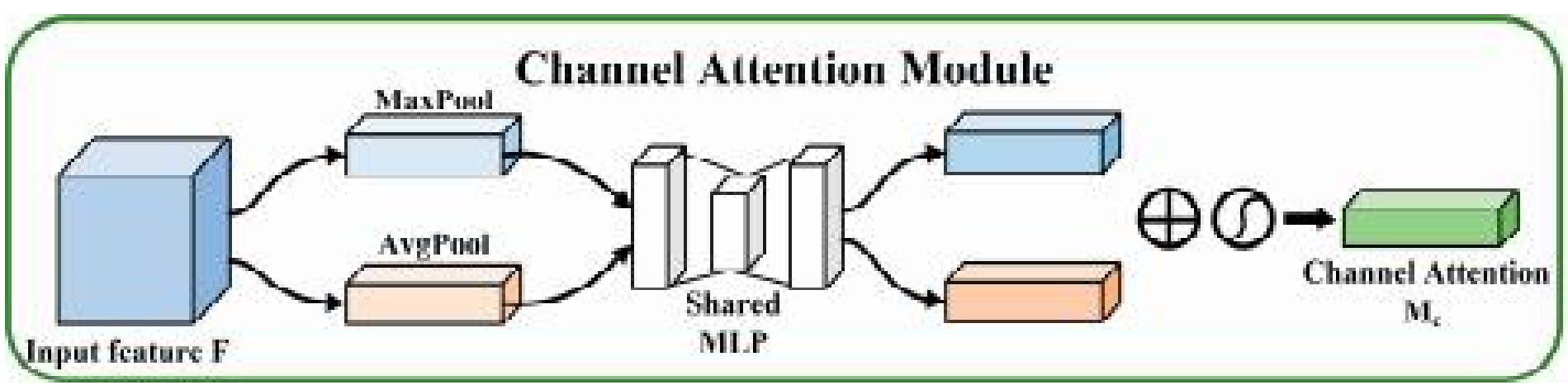

**Figure 3.** Schematic diagram of the channel attention module.

For a given input feature map $F \in R^{C*H*W}$, where $C$, $H$, and $W$ denote the number of channels, height, and width respectively, global average pooling and global max pooling are applied across the spatial dimensions. This operation transforms the input feature map of size C*H*W into two descriptors of size C*1*1, denoted as AvgPool(F) and MaxPool(F). The average pooling computes the mean value of each channel, while the max pooling captures the maximum value within each channel. The corresponding formulas are provided below:

$$\begin{aligned} AvgPool(F) &= \frac{1}{H \times W} \sum_{h=1}^{H} \sum_{w=1}^{W} F(c,h,w) \\ MaxPool(F) &= \max_{h,w} F(c,h,w) \end{aligned} \tag{6}$$

Secondly, the two descriptor vectors are concatenated and passed into a shared multi-layer perceptron (MLP) block. In this block, the channel dimension is first reduced by a factor of r and then restored to its original size. A ReLU activation function is applied to the intermediate outputs, and the results are summed before being passed through a sigmoid activation function to generate the final channel attention map.

$$M_c(F) = \sigma(MLP(AvgPool(F)) + MLP(MaxPool(F))) \tag{7}$$

The resulting channel attention weights are multiplied with the original feature map through element-wise operations to produce the refined feature representation.

$$F' = M_c(F) \otimes F \tag{8}$$

Here, ⊗ denotes element-wise multiplication.

As shown in Figure 4, the Spatial Attention Module is designed to capture spatial location information within the feature map. It assigns a weight to each spatial position, allowing the model to focus more on regions that contain important information. The detailed procedure is illustrated as follows:

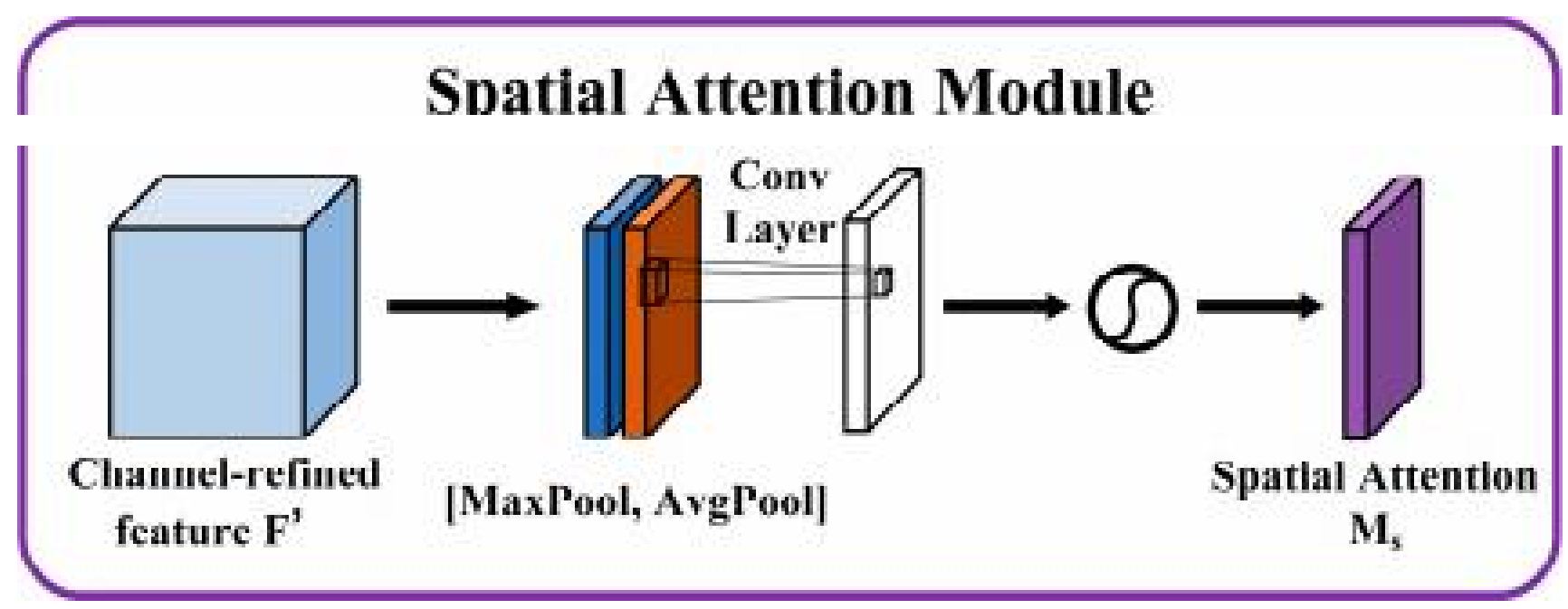

**Figure 4.** Schematic diagram of the spatial attention module.

Channel-wise average pooling and max pooling are separately applied to the input feature map, generating two 2D spatial descriptor maps. These maps are then concatenated along the channel dimension and passed through a convolutional layer—typically with a 7 × 7 kernel—to produce a single-channel feature map. Finally, a sigmoid activation function is applied to generate the spatial attention map. The computation can be formally expressed as follows:

$$M_S(F) = \sigma(Conv([AvgPool(F); MaxPool(F)])) \tag{9}$$

where ; denotes concatenation along the channel dimension, $f^{7\times7}$ represents a convolution with a 7 × 7 kernel, and σ is the sigmoid activation function.

The generated spatial attention weights are applied to the original feature map via element-wise multiplication to obtain the refined feature map.

$$F'' = M_s(F') \otimes F' \tag{10}$$

The entire process can be divided into three steps: 1. The input feature map $F$ is first passed through the Channel Attention Module, resulting in an intermediate feature map $F'$. 2. This intermediate feature map $F'$ is then processed by the Spatial Attention Module to produce the final refined feature map $F''$. 3. The refined feature map $F''$ is subsequently used as input to the next layer in the network, enabling continued forward propagation.

*2.3. Introduction of Discrete Cosine Transform*

The Discrete Cosine Transform (DCT) is a fundamental separable and orthogonal transform widely used in the digital signal processing community. It converts signals from the time or spatial domains into the frequency domain to facilitate energy compaction and feature extraction. Compared to other orthogonal transforms, the DCT employs cosine functions as its basis, which more effectively capture the spatial or temporal correlations in signals such as images and speech. DCT is particularly well-suited for processing multimedia data with high redundancy. The separability and matrix-based representation of DCT form the foundation for understanding classical orthogonal transform algorithms, upon which popular transforms such as the Discrete Cosine Transform (DCT) and Discrete Fourier Transform (DFT) are established.

A generic mathematical representation of a 2D signal transformation can be expressed as follows:

$$T(u,v) = \sum_{x=0}^{M-1}\sum_{y=0}^{N-1} f(x,y)g(x,y,u,v)$$
$$f(x,y) = \sum_{u=0}^{M-1}\sum_{v=0}^{N-1} T(u,v)h(x,y,u,v) \quad (11)$$

Here, $x$ and $y$ represent the spatial coordinates of the original signal, ranging from x, $u = 0, 1, 2, \ldots, M - 1$ and y, $v = 0, 1, 2, \ldots, N - 1$. The function $f(x, y)$ denotes the original image signal, while $T(u, v)$ represents the coefficient matrix after transformation. The functions $g(x, y, u, v)$ and $h(x, y, u, v)$ refer to the forward and inverse transformation kernels, respectively, which define the relationship between the spatial domain and the transformed domain.

The forward and inverse transformation kernels satisfy the following formulas:

$$g(x, y, u, v) = g_1(x, u) \cdot g_2(y, v)$$
$$h(x, y, u, v) = h_1(x, u) \cdot h_2(y, v) \quad (12)$$

This transformation is classified as a separable transform, meaning that the two-dimensional transformation can be decomposed into two independent one-dimensional transforms. Specifically, the transformation is first applied along the x-direction and then along the y-direction, significantly reducing computational complexity. If the functional forms of $g_1$ and $g_2$, as well as $h_1$ and $h_2$, are identical, the transform kernel is said to exhibit symmetry, which further simplifies the transformation process.

A digital image can be represented as a real-valued matrix, where $f(x, y)$ denotes the grayscale matrix of size $M \times N$ Based on the separability of the transform, the transformation process can be succinctly expressed using matrix operations. The specific formula is as follows:

$$\mathbf{F} = \mathbf{PfQ}^T \quad (13)$$

Here, $\boldsymbol{F}$ and $\boldsymbol{f}$ represent the $M \times N$ two-dimensional matrices, corresponding to the transformed coefficient matrix and the original image matrix, respectively. $\boldsymbol{P}$ is an $M \times M$ matrix responsible for the transformation along the row direction, while $\boldsymbol{Q}$ is an $N \times N$ matrix responsible for the transformation along the column direction. The specific expansion is given by the following formula:

$$F(u,v) = \sum_{x=0}^{M-1}\sum_{y=0}^{N-1} P(u,x)f(x,y)Q(v,y) \quad (14)$$

Here, $F(u, v)$ represents the transformed coefficients, while $P(u, x)$ and $Q(v, y)$ denote the elements of the row and column transformation matrices, respectively. Meanwhile, $u = 0, 1, 2, \ldots, M - 1$ and $v = 0, 1, 2, \ldots, N - 1$.

### *2.4. Feature Extraction Layer*

The feature extraction layer in the forgery detection model plays an indispensable role in extracting spatial and frequency domain forgery trace features from processed video frame data. For spatial feature extraction, the model primarily relies on pretrained ResNet50 architecture to extract features from the video frame sequences. ResNet50 employs multiple convolutional layers to capture intra-frame local textures, spatial patterns, and other relevant information. To mitigate overfitting and accelerate the training process, some layers are frozen during training.

Taking better advantage of residual learning, Residual Networks (ResNet) are deep convolutional neural networks that use skip connections to help the network learn the resid-ual (i.e., the difference) between the input and output, rather than learning the mapping directly. This design facilitates the flow of information through the network, significantly alleviating the training challenges of very deep networks and enabling the construction of deeper architectures. The overall architecture of the ResNet50 network is illustrated in Figure 5:

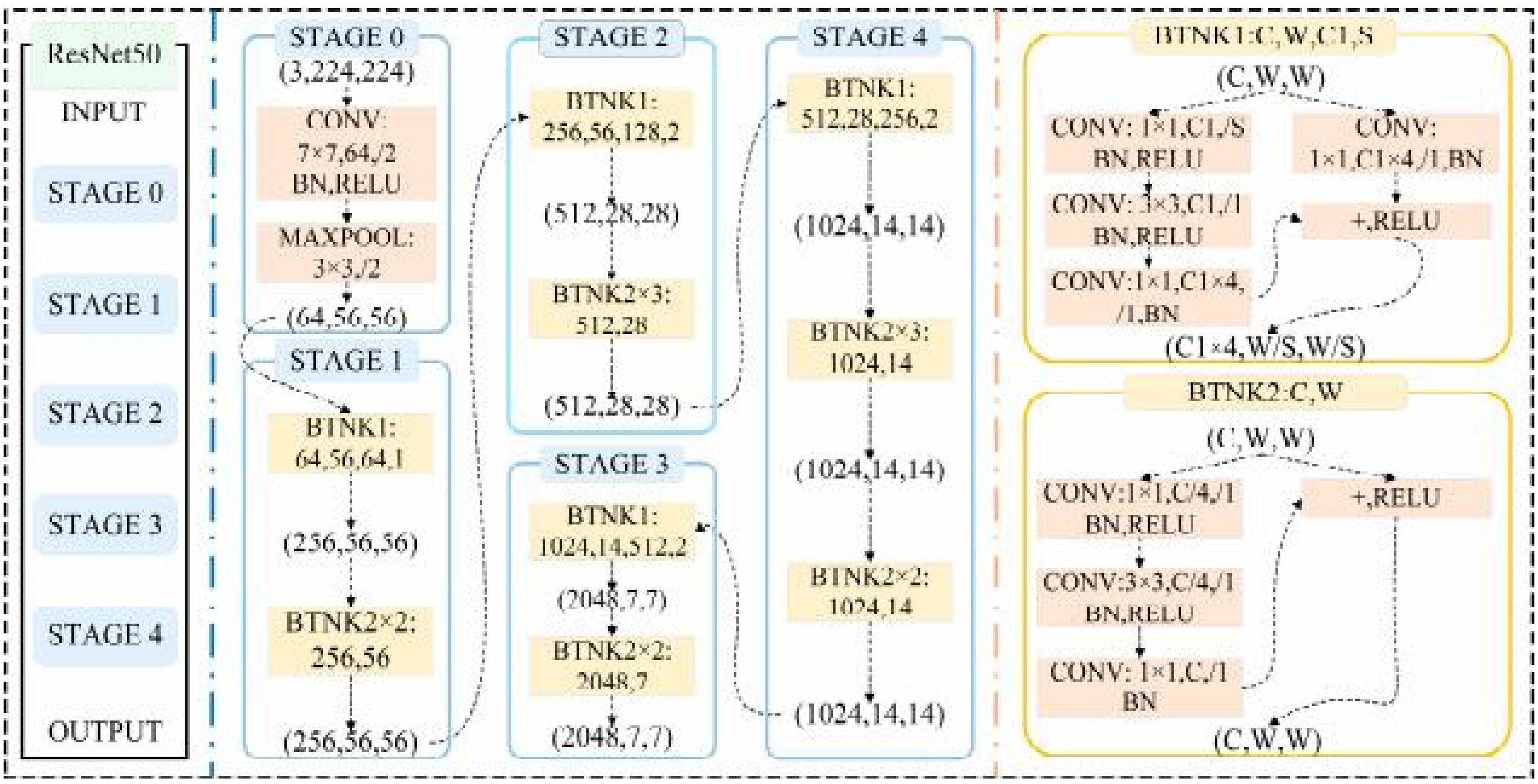

**Figure 5.** Schematic of the ResNet50 Network (BTNK means Bottleneck).

In the frequency domain, the Discrete Cosine Transform (DCT) is also applied to extract frequency domain features, aiding the model in detecting subtle forgery traces. The spatial and frequency domain features are then fused, and the resulting spatial–frequency features are fed into a BiLSTM module to effectively model the temporal dependencies among video frame sequences. The output of the BiLSTM is subsequently passed into a multi-head attention mechanism, which performs multi-head scaled dot-product attention. This mechanism assigns different weights to the corresponding features by estimating their relevance to potential forgery traces, thereby enhancing model sensitivity and ultimately improving detection accuracy.

The spatial and frequency feature extraction blocks are illustrated in Figures 6 and 7, respectively:

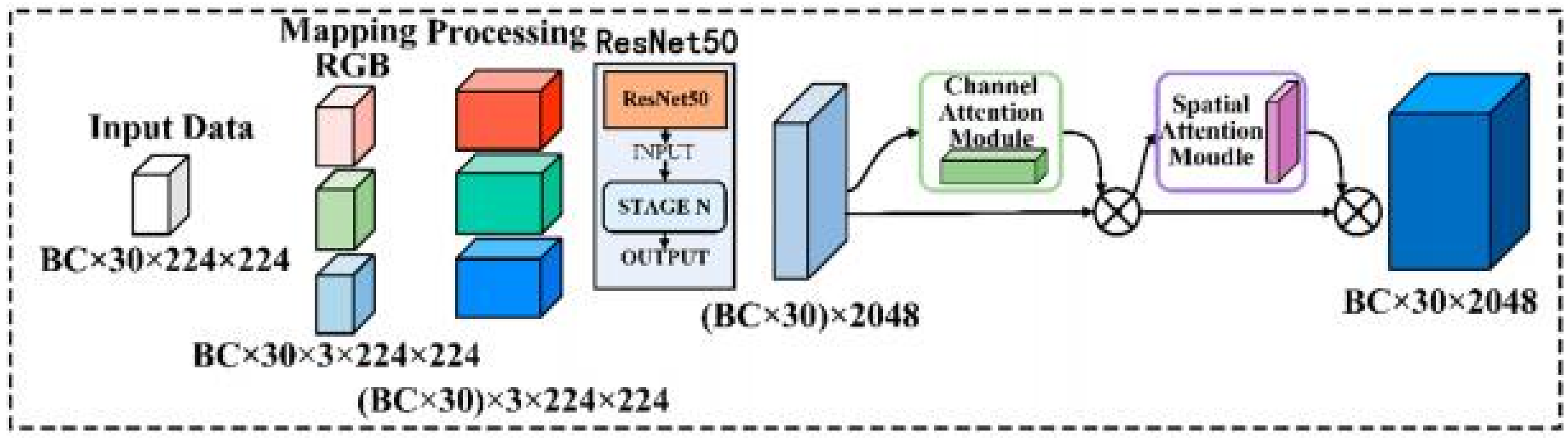

**Figure 6.** Spatial feature extraction block.

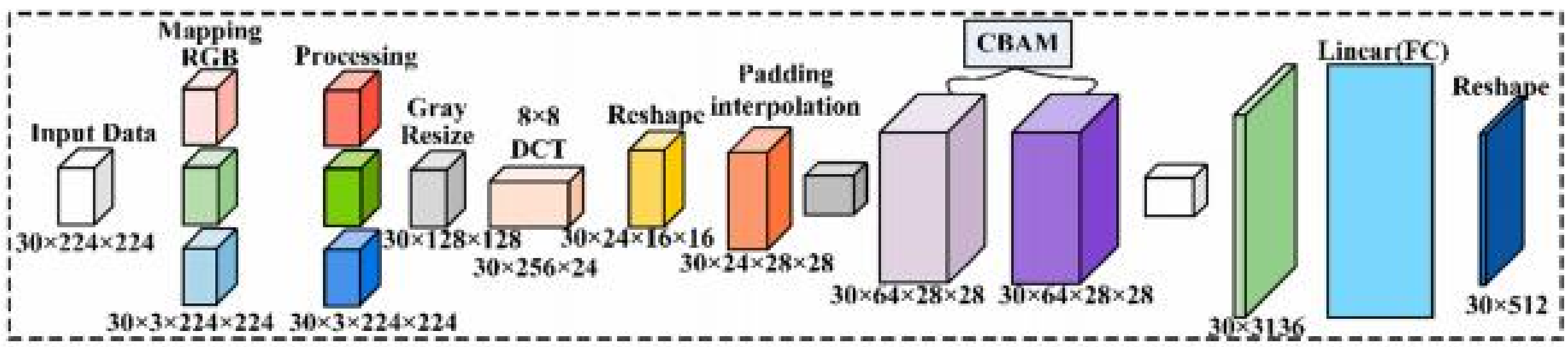

**Figure 7.** Frequency domain feature extraction block.

*2.5. Evaluation Metrics*

Our detection task is formulated as a binary classification problem, where forged (fake) videos are treated as the positive class, and authentic (real) videos as the negative class. To rigorously evaluate the performance of the proposed model, we adopt four widely used metrics: accuracy, precision, recall, and F1 score. Each of these metrics provides unique insight into the model's effectiveness in detecting video forgeries.

Accuracy measures the overall proportion of correctly classified instances from both classes. While it offers a general indication of model performance, its interpretability diminishes under class imbalance—especially when forged content is relatively rare—thus necessitating the use of more discriminative metrics.

Precision refers to the proportion of correctly identified fake videos among all instances predicted as fake. High precision is particularly critical in high-stakes domains such as legal forensics or news verification, where falsely labeling authentic content as forged (i.e., false positives) may result in reputational damage or the invalidation of evidence.

Recall, also known as sensitivity, quantifies the model's ability to correctly identify actual forged content. This metric is particularly important in security-critical scenarios, where false negatives—i.e., failing to detect manipulated media—may lead to the unimpeded spread of misinformation or malicious content.

The F1 score, defined as the harmonic mean of precision and recall, offers a balanced evaluation that reflects the trade-off between Type I and Type II errors. It is particularly informative when assessing models on imbalanced datasets, which are common in real-world forgery detection scenarios.

During the validation process, we sequentially input the test videos into the prediction model. The model then produces preliminary predictions for each video. After aggregating these results, we derive the final prediction and compute the corresponding evaluation metrics.

In addition to the aforementioned standard metrics, we also incorporate several task-specific evaluation indicators to further assess the model's practical effectiveness in real-world scenarios. False Acceptance Rate (FAR) measures the proportion of authentic videos that are incorrectly classified as forged. As a complementary measure to precision, FAR provides direct insight into the model's tendency to raise false alarms. Minimizing FAR is essential in scenarios where false accusations or unnecessary investigations may incur social or legal consequences. Detection Efficiency (DE) reflects the processing speed of the model, typically quantified as the number of videos (or frames) processed per second. A high DE value indicates that the model can operate in real-time or near real-time conditions, which is crucial for time-sensitive applications such as live content monitoring or online platform moderation. Noise Robustness evaluates the model's stability and reliability under various types of input perturbations, such as Gaussian noise, compression artifacts, or adversarial manipulations. A robust model should maintain high detection performance even when video quality is degraded, ensuring its applicability in low-quality or tampered environments commonly found in the wild.

To quantitatively evaluate the model's noise robustness, we followed a controlled experimental design inspired by the methodology in [29]. In consideration of both computational efficiency and runtime constraints, we employed the finite population correction formula (see Equation (20)) to determine an appropriate sample size. The formula is defined as

$$n = \frac{Z^2 \cdot P(1-P) \cdot N}{E^2(N-1) + Z^2 \cdot P(1-P)} \tag{15}$$

where $N$ represents the total population size (in our case, 4297 videos), $n$ is the estimated sample size, $Z$ is the z-score corresponding to the desired confidence level (1.96 for 95% confidence), $P$ is the estimated population proportion (set to 0.5 for maximum variability), and EEE is the margin of error (set to 0.05).

Using this formula, we selected a representative subset of 353 videos through stratified random sampling to maintain the original class distribution. The final sample consisted of 162 forged videos and 191 authentic videos.

It is worth noting that, to simulate noisy real-world environments, we applied additive Gaussian noise at five different standard deviation (SD) levels: 0.01, 0.1, 1, 10, and 100. These levels were chosen to span a wide range of noise intensities, from subtle to severe degradation. The proposed model was then evaluated on each of the five noise-augmented datasets, and the accuracy on the validation set was recorded for each level. By observing the performance degradation trend under increasing noise levels, we were able to assess the model's resilience to input perturbations and thereby gain a deeper understanding of its robustness in practical deployment scenarios.

### 2.6. Metrics of Feasibility

To assess the practical feasibility of the proposed model for real-world deployment, we further evaluated three system-level performance indicators: inference latency, frames per second (FPS), and GPU memory consumption. These metrics provide valuable insight into the model's computational efficiency and suitability for deployment in time-sensitive or resource-constrained environments.

Inference latency refers to the average time the model takes to process a single video input and generate a prediction. Lower latency is crucial in real-time or near-real-time applications—such as live video surveillance or content moderation—where immediate response is required. To ensure the reliability of our latency measurements, we conducted 100 independent inference runs on our video dataset (The frame rate is 30 frames) and computed the average inference time across all trials. This approach mitigates the impact of system fluctuations and provides a stable estimation of the model's typical response time. The final reported latency represents the mean value within a measured range, offering a realistic reflection of expected performance in deployment scenarios.

Frames per second (FPS) quantifies the throughput of the model, indicating how many video frames can be processed per second. Higher FPS values reflect better scalability and efficiency, particularly in scenarios involving long or high-resolution videos. A model with high accuracy but low FPS may not be viable for large-scale or real-time systems.

GPU memory usage (i.e., VRAM consumption) directly affects the deployability of the model on different hardware configurations. Models with high memory footprints may not be suitable for edge devices or shared server environments where resources are limited. Therefore, measuring the memory efficiency of each model helps determine its compatibility with a broader range of deployment platforms.

All experiments were conducted on a system equipped with an NVIDIA RTX 4090 GPU, running CUDA version 11.8.

### 2.7. Statistical Methods

To assess whether the performance difference between our proposed model and the baseline model (CNN-LSTM-Attention) is statistically significant in terms of accuracy and precision, we adopted the 5 × 2 cross-validation paired *t*-test introduced by Dietterich [30]. Specifically, we used five distinct random seeds (0, 42, 123, 2003, and 23,541) to generate five random splits of the original dataset, each split comprising 50% for training and 50% for testing. In each iteration, both models were trained on the training set and evaluated on the corresponding test set, producing performance scores $P_A$ and $P_B$. The training and test sets were then swapped, and the models were re-evaluated to obtain a second set of scores. The differences in performance for each run were subsequently calculated and used in the statistical analysis, The relevant formulas are as follows:

$$P^{(1)} = P_A^{(1)} - P_B^{(1)} \quad (16)$$

$$P^{(2)} = P_A^{(2)} - P_B^{(2)} \quad (17)$$

Then we calculate the estimated mean $\overline{P}$ and variance of the index differences $S2$, and the relevant formulas are as follows:

$$\overline{P} = \frac{P^{(1)} + P^{(2)}}{2} \quad (18)$$

$$S2 = (P^{(1)} - \overline{P})^2 + (P^{(2)} - \overline{P})^2 \quad (19)$$

Next, we calculate the variance across the five iterations and use it to compute the t-statistic as follows:

$$t = \frac{P_1^{(1)}}{\sqrt{(1/5)\sum_{i=1}^{5} s_i^2}} \quad (20)$$

Here, $P_1^{(1)}$ denotes $P^{(1)}$ from the first iteration. We assume that it approximately follows a t-distribution with 5 degrees of freedom and test for a statistically significant difference between the two models at a 95% confidence level. The relevant results are presented in Section 3.5: Experimental Results of Statistics.

## 3. Results and Discussion

### 3.1. Experimental Results

Since the overall model employed an early stopping mechanism with a predefined patience setting, the training process terminated early after the 36 epochs. As training progressed, the loss values of the ResNet-LSTM-Attention video forgery detection model on the training set are shown in Figure 8below.

During the 36 training epochs, depending on the selected evaluation metric, the model retained the model weights from the epoch with the highest accuracy (other criteria can also be chosen based on specific requirements). The performance metrics on both the training and validation sets for that epoch are shown in Figure 9below.

Based on the performance curves of the model in the video forgery detection task, it can be found that the model shows good performance for training and validation. As shown in the figure, the model reaches an accuracy of 91.74% and a precision of 95.75% at the optimum point where the performance can effectively fit the trends of the performance metric on both the training and validation data. According to the test data from the datasets, the model of ResNet-LSTM-Attention can successfully make video forgery predictions between real videos and fake videos and get a reasonable test result. According to the results of test accuracy for two datasets VFDD2.1 and DVF, the model has good predictive capability and is of practical significance.

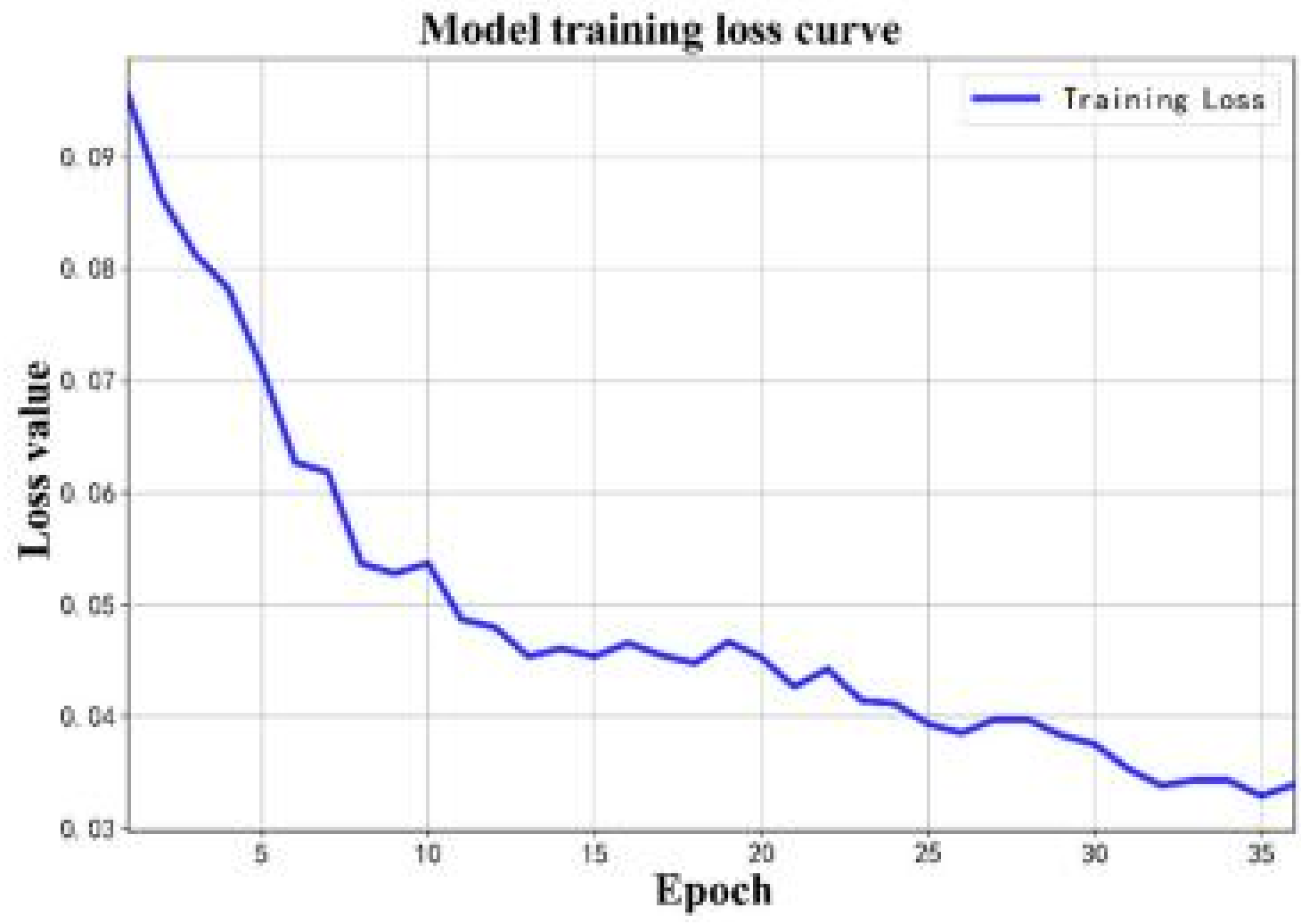

**Figure 8.** The variation curve of our model's loss on the training set.

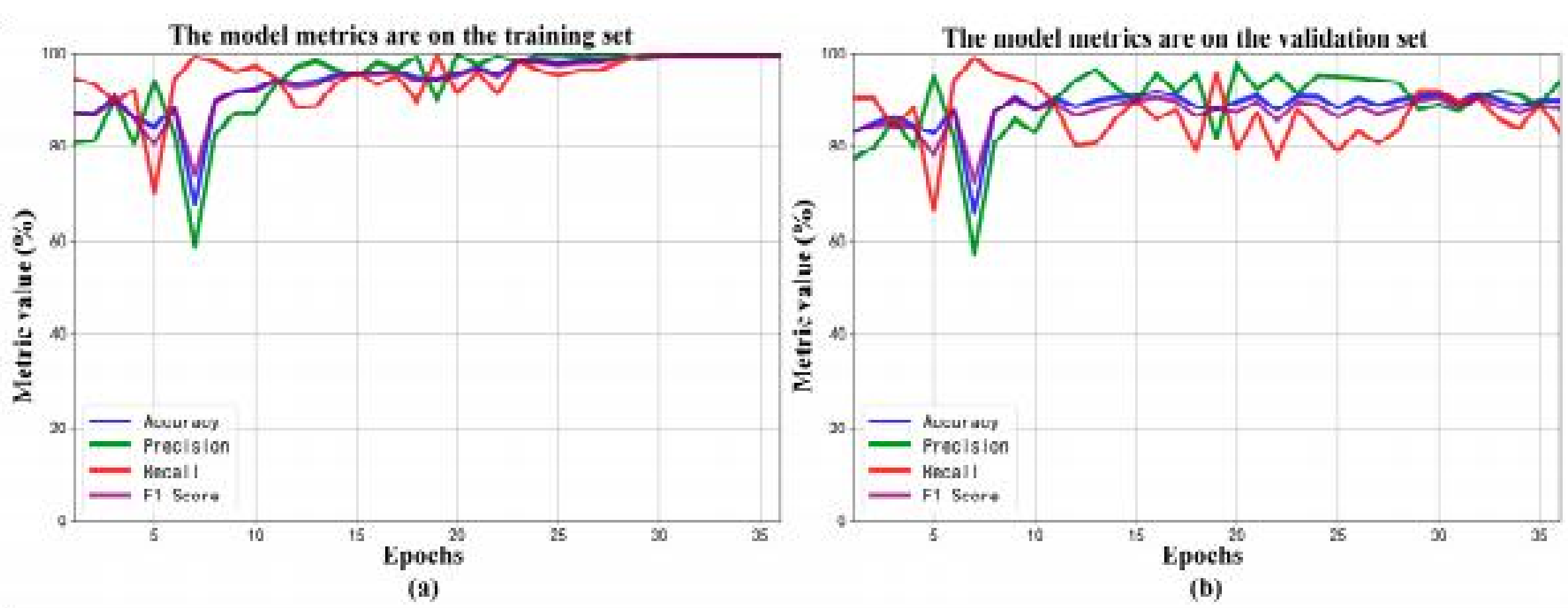

**Figure 9.** Model performance plots for training set (**a**) and validation set (**b**).

*3.2. Comparative Experiment Results*

To verify the effectiveness of the proposed model, comparative experiments were conducted with three configurations: EfficientFormer, ResNet50-RNN-Attention, CNN-LSTM-Attention, and ViTranSPAD [31]. Among them, CNN-LSTM-Attention serves as the baseline model, which utilizes a traditional convolutional backbone and sequential learning components. ResNet50-RNN-Attention is included to assess the impact of different temporal modeling units (RNN vs. LSTM) on the performance of deep spatial–temporal architectures. In contrast, EfficientFormer is a lightweight transformer-based model chosen to explore the trade-off between computational efficiency and predictive accuracy. In addition, ViTranSPAD (referred to as ViTranSP blow) is included as an advanced reference model to benchmark performance under transformer-based paradigms. ViTransPAD introduces a Multi-scale Multi-Head Self-Attention (MsMHSA) mechanism and integrates convolutional token embedding and projection modules to simultaneously capture local spatial features and long-range temporal dependencies. The accuracy, precision, recall, and F1 scores are shown in (a), (b), (c), and (d) in Figure 10, respectively (since each model is trained on different rounds, the highest rounds are shown and the training is stopped at the specified position after the end).

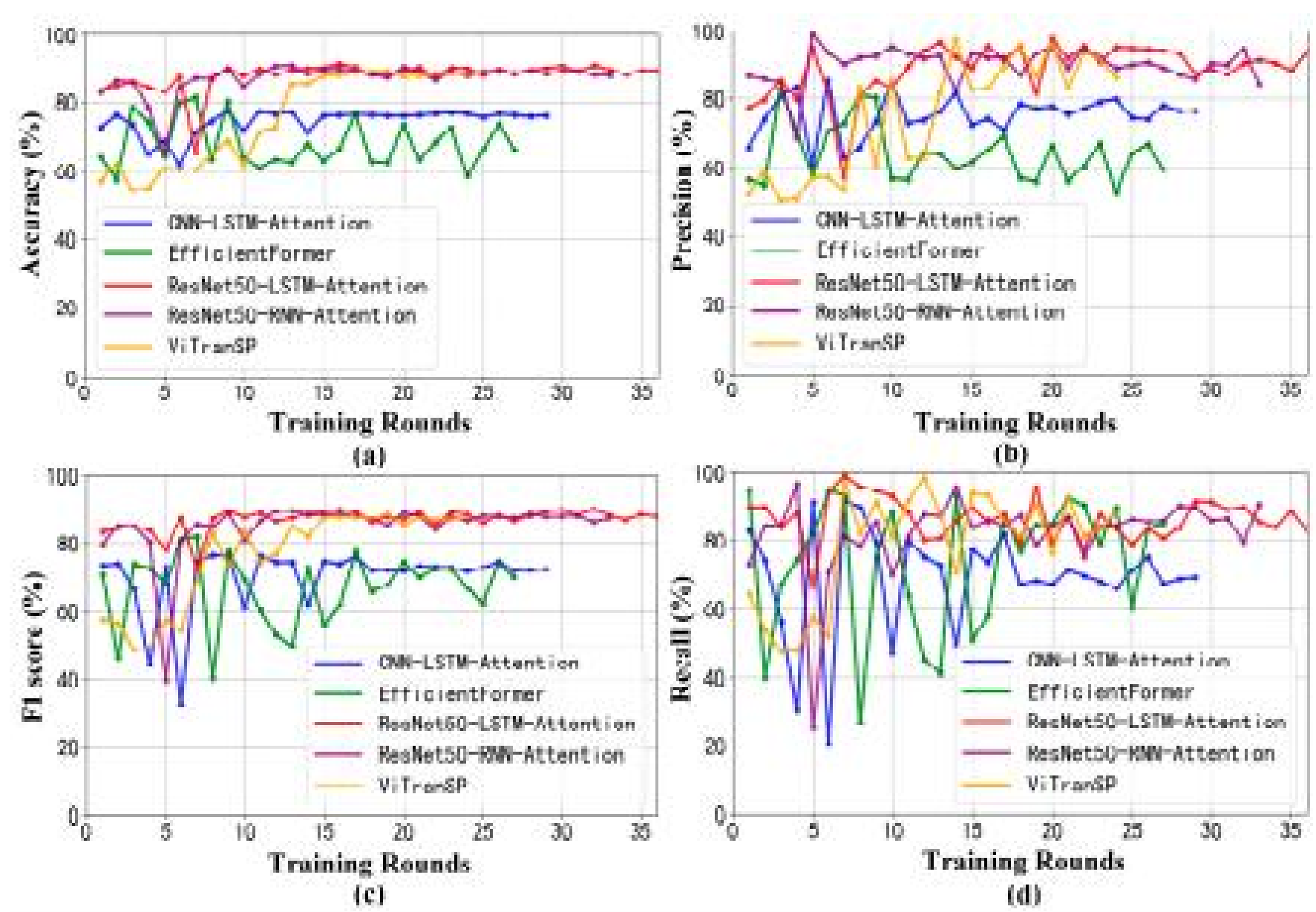

**Figure 10.** Accuracy comparison chart (**a**), precision comparison chart (**b**), F1 score chart (**c**), recall score chart (**d**) of each model.

The comparative experimental results of the VFDD2.1 and DVF datasets (with the set of data with the highest accuracy rate for verification as the standard) are shown in Table 1.

**Table 1.** Comparative experiment results.

| Model Name | Accuracy | Precision | Recall | F1 Score | FAR | DE (Day) |
|---|---|---|---|---|---|---|
| EfficientFormer | 83.28% | 73.16% | 93.40% | 84.01% | 28.96% | 77.02 |
| ResNet50-RNN-Attention | 91.28% | 92.99% | 87.56% | 90.19% | 5.58% | 39.62 |
| CNN-LSTM-Attention | 77.56% | 73.54% | 79.70% | 76.50% | 24.23% | 39.26 |
| ResNet50-LSTM-Attention | 91.74% | 95.75% | 85.79% | 90.50% | 3.22% | 86.29 |
| ViTranSP | 89.77% | 87.14% | 91.12% | 89.09% | 11.38% | 89.42 |

In the comparative experiments, based on the results from the validation set, the ResNet50-LSTM-Attention model exhibited the best overall performance across all evaluated metrics. It achieved the highest accuracy (91.74%), precision (95.75%), and F1 score (90.50%), as well as the lowest False Acceptance Rate (3.22%), underscoring its strong ability to distinguish forged content from real video inputs. It also demonstrated a high Detection Efficiency (DE) score of 86.29, making it suitable for practical applications that demand both effectiveness and reliability.

However, one limitation of the ResNet50-LSTM-Attention model is its training time, which amounted to 1 day, 1 h, 32 min and 35 s. This relatively long training duration may present constraints in scenarios requiring frequent model updates or large-scale experimentation. Additionally, although its DE score is high, it is slightly lower than that of ViTranSP, which recorded the highest DE (89.42) among all models, suggesting that ViTranSP offers greater computational efficiency.

Among the baseline models, CNN-LSTM-Attention showed the weakest performance across most metrics, with an accuracy of 77.56% and a relatively high FAR of 24.23%, indicating limited reliability. EfficientFormer, though more efficient in terms of GPU memory usage, demonstrated only moderate accuracy (83.28%) and suffered from the highest FAR (28.96%), limiting its applicability in high-precision scenarios. ResNet50-RNN-Attention, while slightly inferior to our proposed model, still performed competitively, with an accuracy of 91.28% and a relatively low FAR of 5.58%, illustrating the benefit of deep convolutional spatial encoding combined with recurrent temporal modeling. ViTranSP, as transformer-based architecture integrating convolutional operations and multi-scale spatio-temporal attention, offered well-balanced performance with accuracy of 89.77%, recall of 91.12%, and an F1 score of 89.09%. While its precision (87.14%) was lower than that of our proposed model, it achieved the highest detection efficiency, making it especially suitable for real-time or resource-constrained deployments. Overall, these results confirm that ResNet50-LSTM-Attention not only excels in classification accuracy but also offers significant advantages in practical deployment due to its high detection efficiency. Nonetheless, ViTranSP serves as a strong complementary solution, especially when computational efficiency is a higher priority. The comparative analysis underscores the importance of selecting a model based not only on accuracy but also on practical deployment considerations such as training cost and runtime efficiency.

To further assess the robustness of different models under noisy conditions, we introduced varying levels of Gaussian noise by adjusting the resampling rate from $10^{-2}$ to $10^{2}$, as illustrated in Figure 11. Among all models, ViTranSP and EfficientFormer exhibited the strongest noise robustness, maintaining relatively stable accuracy curves across the entire range of resampling rates. ViTranSP, with its convolution-augmented transformer structure, showed a strong ability to resist performance degradation under noise corruption, while EfficientFormer also demonstrated excellent resilience even at extreme noise levels, with only slight performance declines. These results suggest that both models are highly tolerant of input degradation and noise interference, making them suitable for deployment in real-world environments where signal quality may vary. The proposed ResNet50-LSTM-Attention model ranked second in terms of noise resistance. Although its accuracy curve showed more fluctuation than that of EfficientFormer, it remained consistently above the baseline CNN-LSTM-Attention and outperformed ResNet50-RNN-Attention under most noise conditions. Notably, ResNet50-LSTM-Attention demonstrated a strong sensitivity to low levels of noise: as the resample rate increased slightly (from $10^{-2}$ to $10^{-1}$), the model's accuracy improved significantly, showing the largest accuracy gain among all models in this low-noise region. This suggests that minor perturbations in input may help the model generalize better by preventing overfitting, further highlighting its adaptability. In contrast, the baseline CNN-LSTM-Attention model experienced severe performance degradation under noisy conditions, confirming its limited robustness.

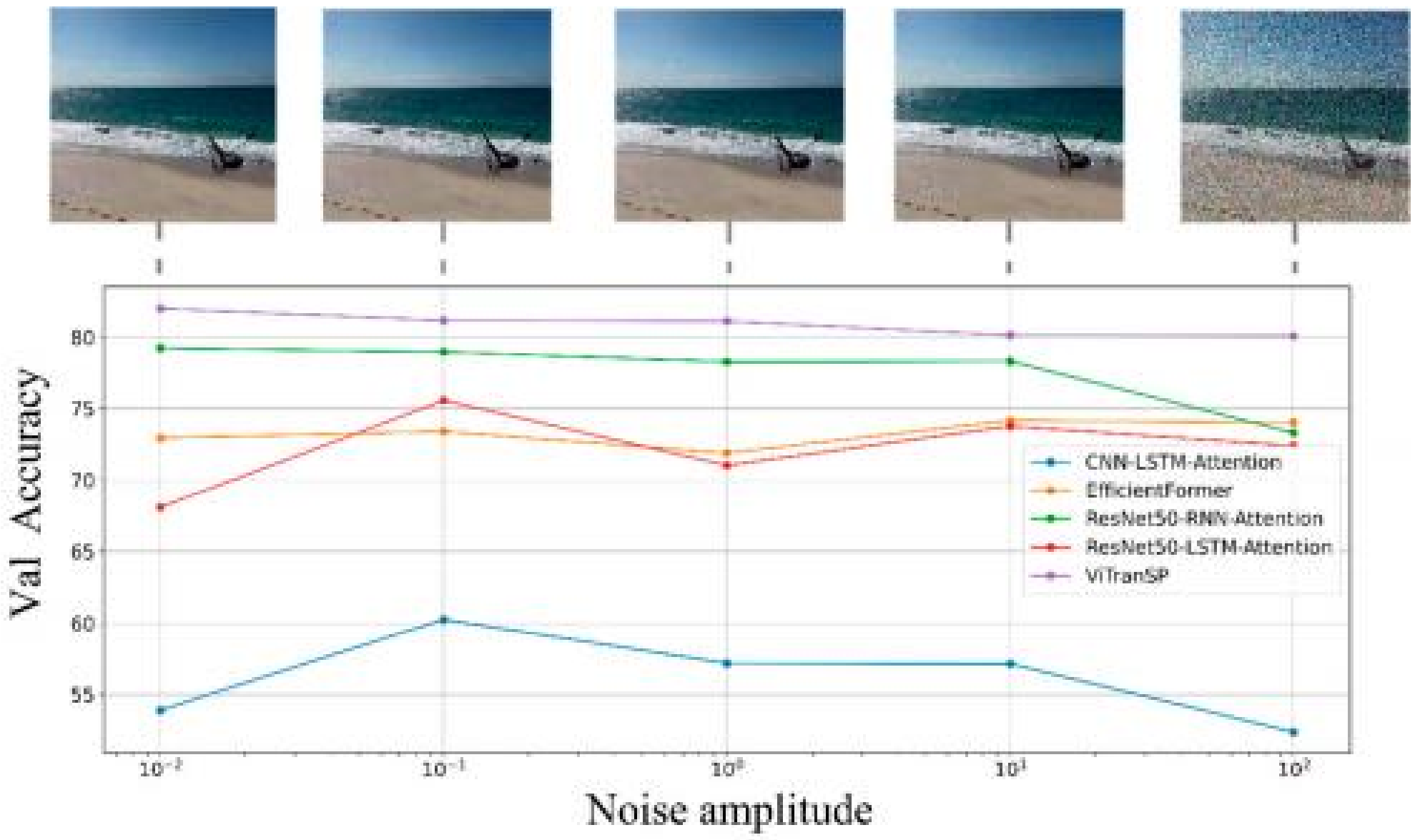

**Figure 11.** The accuracy change graph for the model validation set under the five noise levels.

In summary, although ViTranSP and EfficientFormer achieved the best overall resistance to noise, ResNet50-LSTM-Attention still preserved high accuracy across varying noise levels and demonstrated adaptability to slight perturbations.

### *3.3. Ablation Experiment Results and Discussion*

To evaluate the contribution of each component in the proposed algorithm, the overall ablation experiments are divided into $O_1$ without DCT feature extraction, $O_2$ without BiLSTM, and $O_3$ without the multi-head attention mechanism. The ablation video forgery detection model, as it progresses through different epochs, shows changes in accuracy, precision, recall, and F1 score, which are illustrated in Figure 12a–d. (Since each model was trained for a different number of epochs, the highest number of epochs is displayed, with training stopping at the specified position):

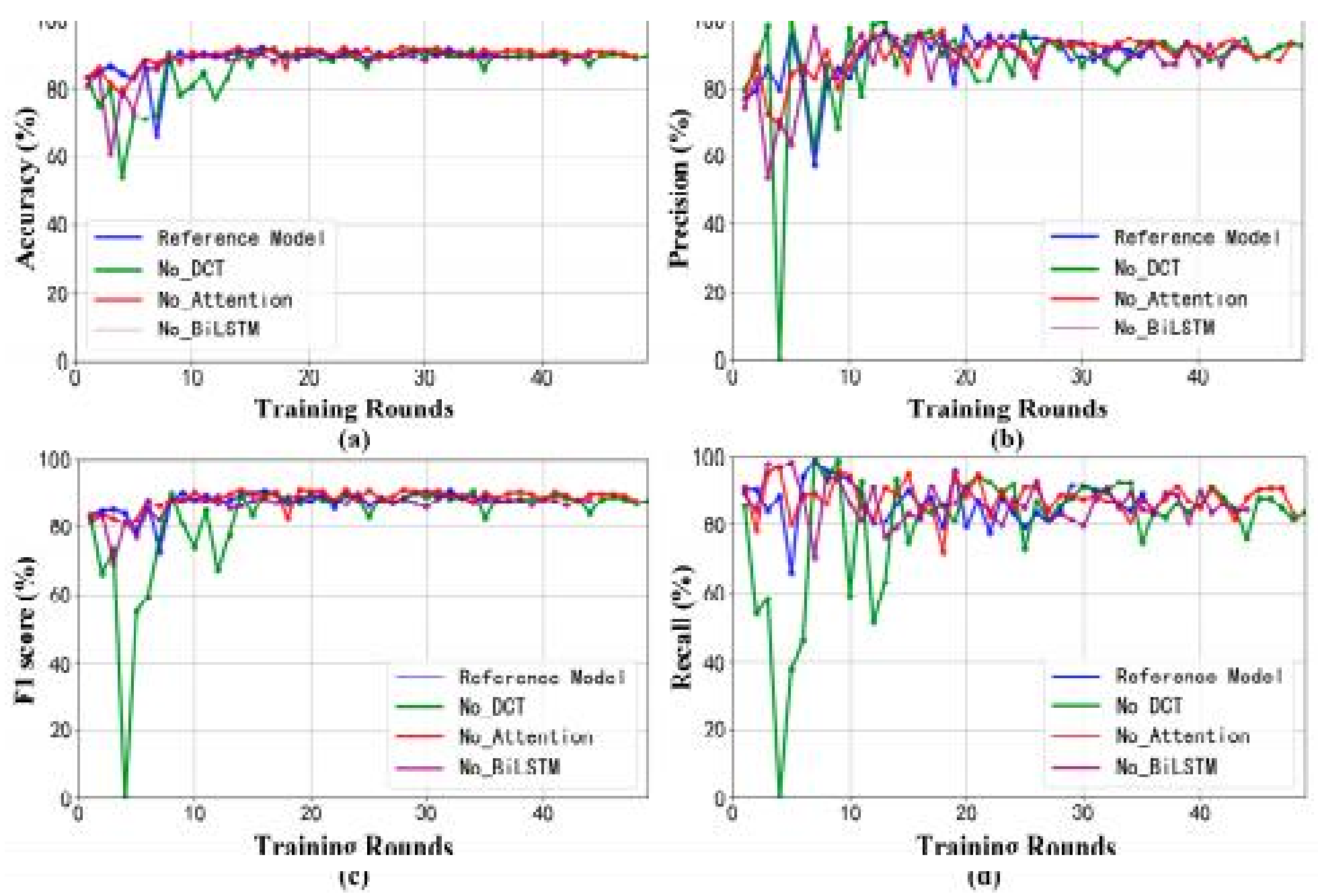

**Figure 12.** Model accuracy comparison (**a**), precision comparison (**b**), F1 score (**c**), and recall score (**d**) in ablation experiment.

The ablation experimental results of the VFDD2.1 and DVF datasets (the set of data with the highest accuracy of the validation data is taken as the standard) are shown in Table 2.

**Table 2.** Ablation experiment results.

| Model | Accuracy | Precision | Recall | F1 Score | Used Time (Day-Hour:Minute:Second) |
|---|---|---|---|---|---|
| No_DCT | 91.28% | 92.53% | 88.07% | 90.24% | 0-19:48:02 |
| No_Attention | 91.86% | 93.09% | 88.83% | 90.91% | 3-07:56:54 |
| No_BiLSTM | 90.93% | 92.25% | 87.56% | 89.84% | 2-02:10:38 |
| Reference | 91.74% | 95.75% | 85.79% | 90.50% | 1-01:32:35 |

Based on the ablation experiment data from validation set of the analysis, we conclude that the model has better performance for the whole training task on the video forgery detection task with 91.74% and performs better than the other three ablation models, 0.12 lower than the “No Attention” model’s 91.86%. However, its inference time is only 1 day, 1 h, and 32 min, which is just one-third of the “No Attention” model’s 3 days, 7 h, and 56 min—saving approximately 54 h. This demonstrates a significant improvement in time efficiency, despite the minimal accuracy gap. With classification accuracies of 91.28% (0.46% higher) and 90.93% (0.81% higher), it shows that the baseline model is quite stable and reliable in classification accuracies compared with the "No DCT" model and "No BiLSTM" model.

Furthermore, by the ablation experiments, removing DCT only led to accuracy down by 0.46% but substantial speedup, whereas removing BiLSTM had an accuracy impact of 0.81% and substantial slow down. Thus, we conclude that the baseline model, ResNet50-LSTM-Attention, maintains reasonable accuracy. Its slight loss in accuracy compared to the "No Attention" model is offset by a much shorter inference time, indicating better real-time performance. It satisfies practical requirements for video forgery detecting to a certain extent. The good performance and feasible computational efficiency lead to a trustworthy solution for video forgery detection tasks.

### *3.4. Experimental Results of Feasibility*

The results of feasibility assessment by different models are presented in Table 3

**Table 3.** Feasibility evaluation metrics across different models.

| Model Name | Latency (ms) | FPS | GPU Memory |
|---|---|---|---|
| EfficientFormer | 107.35 ± 0.75 | 9.31 | 1963 MB |
| ResNet50-RNN-Attention | 56.52 ± 2.04 | 17.69 | 1791 MB |
| CNN-LSTM-Attention | 14.69 ± 0.10 | 68.08 | 5517 MB |
| ResNet50-LSTM-Attention | 21.51 ± 0.15 | 46.49 | 3377 MB |
| ViTranSP | 19.74 ± 0.38 | 50.66 | 2074 MB |

Our proposed model achieves a favorable trade-off between latency and throughput, with an average inference latency of 21.51 ± 0.15 ms, FPS of 46.49, and GPU memory usage of 3377 MB. This suggests that the model is both responsive and efficient enough for real-time applications, while maintaining a moderate memory footprint compatible with most modern GPUs.

Among the comparison models, EfficientFormer exhibits the highest latency (107.35 ± 0.75 ms) and the lowest FPS (9.31), indicating a relatively slow inference speed. However, it consumes low GPU memory (1963 MB), suggesting its potential suitability for deployment in memory-constrained environments. ViTranSP, another strong baseline, demonstrates competitive efficiency, with a low latency of 19.74 ± 0.38 ms and high throughput (50.66 FPS), while keeping GPU usage relatively low (2074 MB). Its performance is close to our proposed model and reflects the advantages of incorporating lightweight vision transformers in terms of both speed and memory economy.

In contrast, CNN-LSTM-Attention achieves the highest FPS (68.08) and the lowest latency (14.69 ± 0.10 ms), but at the cost of substantially higher GPU memory consumption (5517 MB), which may limit its applicability in hardware-limited scenarios.

Overall, while our model does not offer the most competitive GPU memory efficiency, and its inference speed lags slightly behind that of the best-performing model ViTranSP, it achieves a well-balanced trade-off among latency, throughput, and resource consumption. This balance makes it particularly suitable for deployment in scenarios with reasonable resource constraints, where both responsiveness and computational feasibility are required.

### *3.5. Experimental Results of Statistics*

The results of the significance test (with a confidence interval of 95%) of our module and baseline are shown in Table 4.

To further validate whether the performance difference between the proposed ResNet50-LSTM-Attention model and the baseline CNN-LSTM-Attention model is sta-tistically significant, we conducted a 5 × 2 cross-validation paired *t*-test following the procedure outlined in Section 2.7. The results of the *t*-test for accuracy and precision are shown in Table 4. For accuracy, the computed t-value was 13.088 with a corresponding *p*-value of 0.000, indicating a statistically significant difference at the 95% confidence level. This confirms that our proposed model consistently outperforms the baseline model in terms of classification correctness. In contrast, the *t*-test for precision yielded a t-value of 2.332 and a *p*-value of 0.067, which exceeds the standard significance threshold of 0.05. Therefore, the improvement in precision, while numerically higher, is not statistically significant under the given test settings. One possible explanation is that both models tend to be relatively conservative in their predictions of positive (forged) samples, resulting in similar levels of false positives. Additionally, the relatively small size of the test samples in each fold (due to 50% splits) and the inherently high variance in precision in imbalanced classification tasks may have contributed to this result. The relevant results are presented in Figure 13.

**Table 4.** Statistical comparison of accuracy and precision between our module and baseline using 5 × 2 CV paired *t*-Test.

| | In Accuracy | In Precision |
|---|---|---|
| *t*-test | 13.088 | 2.332 |
| *p*-value | 0.000 | 0.067 |

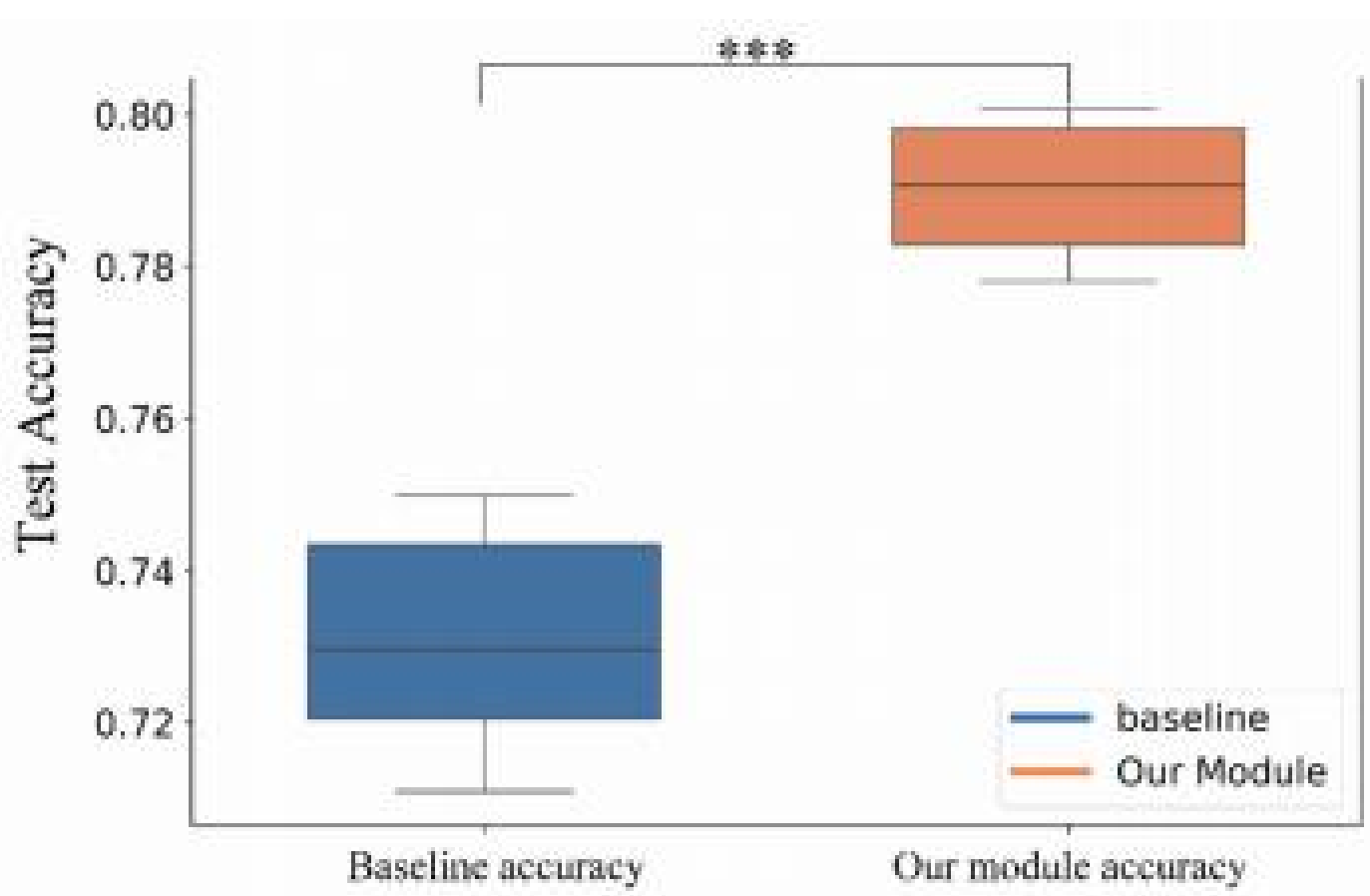

**Figure 13.** Boxplot comparison of accuracy between baseline and proposed model (*** mean $p < 0.001$).

Overall, these findings reinforce the idea that the proposed model's improvement in accuracy is not only meaningful but statistically reliable, while the difference in precision, though present, requires further investigation with possibly larger or more diverse datasets to achieve statistical significance.

## 4. Conclusions

This study develops a ResNet-LSTM-Attention model for video forgery detection using Python3.10 and thoroughly verifies its effectiveness in detecting video tampering. Through observation of the trend in the model's loss value, accuracy, precision, recall, and F1 score on the validation set in the training period and association with the visual results and statistics results, this paper comprehensively discusses the model's performance from the perspectives of qualitative analysis and quantitative analysis. Experimental results show that the model, built with a spatiality–frequency fusion strategy and an attention mechanism, achieves 91.74% accuracy and 95.75% precision on the VFDD2.1 and DVF datasets. It effectively handles various tampering scenarios, including frame deletion, replacement, insertion, and AIGC-based forgery.

Through ablation experiments, this paper further reveals the important contribution of the DCT, BiLSTM, and multi-head attention mechanisms to the model's performance and finds that the baseline model achieves an ideal balance between accuracy and inference time. Furthermore, in our comparative experiments, although it did not reach the same level as ViTranSP and EfficientFormer in some indicators, its superiority was still demonstrated.

Overall, this chapter experimentally presents a rigorous process to illustrate good performance of the model in video forgery detection tasks, including data preprocessing, design of model, optimization of hyperparameters, and experimental validation. The following conclusions can be drawn:

1. This paper addresses the limitation of existing video forgery detection methods in deep learning, which mainly focus on single-dimensional features and single tampering types. The proposed ResNet-LSTM-Attention model, which integrates spatiality–frequency features and multi-head attention mechanisms, effectively identifies various tampering types, such as frame deletion, insertion, replacement, and AIGC forgery.
2. Through extensive experiments, the model achieves 91.74% accuracy and 95.75% precision on the VFDD2.1 and DVF datasets. Qualitative and quantitative evaluations confirm the model's outstanding performance.
3. The ablation experiments show that the DCT module, BiLSTM structure, and attention mechanism are crucial for improving the model's performance. The baseline model achieves a good balance between accuracy and inference efficiency.
4. The proposed method is compared with some benchmark models on the classification task. In the classification task, compared to other models, the proposed method is effective and time-efficient, and it is of greater practical use. In future, we will tune the structure of our proposed algorithm in order to make the algorithm more efficient with real-time performances.

Despite the promising performance demonstrated by our proposed model, several limitations should be acknowledged. First, the model has not been tailored for any specific application domain; all experiments were conducted on publicly available datasets containing general, everyday video content. As a result, its effectiveness in domain-specific scenarios (e.g., medical videos, surveillance footage, or industrial inspection) remains untested. In future work, we plan to develop customized forgery detection models that are trained and optimized for such specialized domains, thereby enhancing real-world applicability. Secondly, the current model relies on a relatively large amount of labeled training data to achieve optimal performance. The availability of sufficient training samples is a critical factor in ensuring accuracy, and when data is limited, the model's detection ability may be compromised. To address this limitation, future research will focus on optimizing the network structure and exploring techniques such as few-shot learning or data augmentation, with the goal of maintaining high performance even under low-resource conditions.

**Author Contributions:** Conceptualization, Y.C.; validation, Y.C.; data curation, Z.L.; writing—original draft, Z.L.; formal analysis, Z.L.; writing—review and editing, Z.L.; Supervision, S.H. All authors have read and agreed to the published version of the manuscript.

**Funding:** This work is supported by the National Key Research and Development Program of China (Grant No. 2022YFB3103602).

**Institutional Review Board Statement:** Not applicable.

**Informed Consent Statement:** Not applicable.

**Data Availability Statement:** The DVF dataset that support the findings ofthis study will be available in github at https://github.com/SparkleXFantasy/MM-Det(accessed on 1 May 2025); the VFDD2.1 dataset is available from South China University of Technology. Contact eeyjhu@scut.edu.cn for access or it is also available from the corresponding author upon reasonable request.

**Acknowledgments:** We would like to thank South China University of Technology for providing VFDD2.1 data support, especially Hu Y et al.

**Conflicts of Interest:** No potential conflicts of interest were reported by the authors.